\documentclass{article}

\PassOptionsToPackage{round,semicolon}{natbib}  % citations as (Author, Year; Author, Year), as in the draft
\usepackage[dblblindworkshop,final]{neurips_2026}
\workshoptitle{Second Workshop on Large Language Models for Social Reasoning and Simulation}

\usepackage[utf8]{inputenc}
\usepackage[T1]{fontenc}
\usepackage{amsmath}
\usepackage{amsfonts}
\usepackage{fdsymbol}
\usepackage{booktabs}
\usepackage{graphicx}
\usepackage{xcolor}
\usepackage{wrapfig}
\usepackage{enumitem}
\usepackage{tcolorbox}
\tcbuselibrary{breakable,skins}
\usepackage{array}
\usepackage{url}
\usepackage{needspace}
\usepackage[hidelinks]{hyperref}
\newcommand{\arena}{\textsc{PersuArena}}
\newcommand{\method}{\textsc{Probe}-RPO}
\newcommand{\dense}{\mathrm{dense}}
\newcommand{\Rout}{R^{\mathrm{out}}}

\setlist[itemize]{leftmargin=*,topsep=4pt,itemsep=2pt}

\title{Implicit Beliefs as Scalable Process Rewards for\\Strategic Persuasion}

\author{%
  Minqian Liu\textsuperscript{1}\quad
  Yuexi Shen\textsuperscript{4}\quad
  Heajun An\textsuperscript{1}\quad
  Qi Zhang\textsuperscript{1} \\
  \textbf{Pamela J. Wisniewski\textsuperscript{3}\quad
  Sang Won Lee\textsuperscript{1}\quad
  Jin-Hee Cho\textsuperscript{1}\quad
  Lifu Huang\textsuperscript{2}} \\
  \textsuperscript{1}Virginia Tech\quad
  \textsuperscript{2}UC Davis\quad
  \textsuperscript{3}STIR Lab\quad
  \textsuperscript{4}UC Santa Barbara
}

\begin{document}

\maketitle

\begin{abstract}
Training large language models (LLMs) for multi-turn strategic persuasion faces two coupled obstacles: (i) realistic simulation of psychologically grounded persuadees, and (ii) dense, faithful supervision for credit assignment in strategic dialogues. Outcome-only rewards are sparse and uninformative about \emph{which} conversational moves drove success, while LLM-judge process rewards are expensive and decoupled from the underlying interaction dynamics. We address both challenges with a unified framework. First, we introduce \arena{}, a persuasion simulation environment that produces realistic, non-monotonic trajectories. Second, building directly on this environment, we propose \textbf{P}rocess \textbf{R}ewards via \textbf{O}nline \textbf{B}elief \textbf{E}licitation for Relative Policy Optimization (\method{}), an online RL framework that converts the persuadee's internal state into dense process rewards. At each turn, three introspective probes are read from the frozen persuadee's logits, yielding a multi-dimensional reward signal that is faithfully aligned with persuasion dynamics. With extensive experiments on \arena{} spanning healthcare, financial advising, workplace management, education, and personal relationships, we demonstrate that our \method{} achieves the full acceptance rate of 52.5\%, compared with the 35.8\% for the strongest GRPO ablation and 25.8\% for the best off-the-shelf LLM. We also conduct comprehensive analyses on the generalization across persuadee models and the persuasion dynamics across conversations. Our work offers a promising paradigm that can provide effective process supervision for strategic dialogue.
\end{abstract}

% ============================== 1 INTRODUCTION ===============================
\section{Introduction}

Persuasion, i.e., the capacity to shift another person's beliefs, attitudes, or behavioral intentions through strategic communication, underlies many high-stakes human interactions, from clinical counseling and public-health intervention to negotiation and crisis de-escalation \citep{lewis-etal-2017-deal,persuasion4good,persuasion_survey}. As large language models (LLMs) \citep{achiam2023gpt,touvron2023llama} are increasingly deployed as autonomous conversational agents in these domains, a pressing question emerges: \emph{how can we systematically learn a policy to navigate the complex, multi-turn dynamics of real-world persuasion?}

Two tightly coupled challenges make this question difficult. The first is \textbf{realistic simulation}. Real persuasion targets do not exhibit monolithic resistance; they push back from psychologically distinct mechanisms, e.g., entrenched belief, identity threat, autonomy-driven reactance, or genuine ambivalence, each demanding qualitatively different strategies \citep{krosnick1995attitude,brehm1966theory,oyserman2009identity}. They also evolve dynamically: repeated pressure escalates reactance, and the same argument used too many times triggers inoculation \citep{mcguire1964inducing}. Existing LLM evaluation frameworks largely ignore these dynamics: the simulated persuadee's behavior typically emerges from the LLM's own priors rather than from principled psychological structure, yielding interactions that are unrealistically compliant or arbitrarily obstinate.

The second challenge is \textbf{dense supervision}. Multi-turn persuasion is a kind of long-horizon, strategy-intensive task where outcome-only rewards are uninformative, e.g., a final accept/reject signal does not tell the policy whether success came from skillful diagnosis of the persuadee's underlying concerns or from some random factors. Recent RL recipes for social agents \citep{yu2025sotopia} therefore add dedicated LLM judges that score each utterance after the fact. However, such an approach is architecturally decoupled from the very interaction dynamics the agent must learn, and also expensive as it introduces an additional large model in the training loop.

We observe that these two challenges share a common solution: a simulator built for psychological realism can also serve as a source of dense supervision. If the persuadee itself is injected by principled psychological mechanisms, then it can be read out directly, at the cost of a single forward pass, to produce a dense, faithful reward signal that is structurally tied to the dynamics governing the outcome. With this motivation, we first introduce \arena{} (\S\ref{sec:arena}), a theory-grounded persuasion simulation environment built on a comprehensive taxonomy that operationalizes established psychological theories \citep{petty1986communication,brehm1966theory,mcguire1964inducing,mayer1995integrative} into executable simulation constraints. We curate 120 scenarios spanning healthcare, financial advising, workplace management, education, and personal relationships, with combinatorial coverage of resistance types, motivational orientations, and stake levels, such that \arena{} consists of diverse and realistic persuadee profiles that enable faithful estimation of the persuasion state.

Building on this environment, we then propose \textbf{P}rocess \textbf{R}ewards via \textbf{O}nline \textbf{B}elief \textbf{E}licitation for Relative Policy Optimization (\method{}; \S\ref{sec:method}), an online RL framework that derives dense, multi-dimensional process rewards directly from the persuadee's inner belief-state transitions. After each turn, a two-pass rollout generates the persuadee's visible reply and then implicitly measures the persuasion state through introspective probing. The per-turn reward is computed as the change in the resulting belief state, a dense signal that is faithful to the persuadee's own representation. We optimize the persuader with a critic-free policy update adapted from GRPO \citep{shao2024deepseekmath}, using a composed advantage estimator that combines the outcome-level and process-level signals.

Through extensive experiments, our key findings are: \textbf{(i)} Our \method{} substantially outperforms the strongest outcome-only GRPO baseline by \textbf{16.7 percentage points} in full acceptance rate, which indicates our method can provide high-quality dense supervision, provided that the process reward is weighted moderately (\S\ref{sec:main}). \textbf{(ii)} Our learned policy can transfer to different persuadee models unseen during training, which demonstrates the generalizability of our method (\S\ref{sec:generalization}). \textbf{(iii)} We reveal the underlying patterns of persuasion dynamics for different models (\S\ref{sec:dynamics}), and illustrate that our designed introspective probes are effective indicators of persuasion outcomes (\S\ref{sec:probetraj}). In summary, we make the following contributions:
\begin{itemize}
  \item We introduce \textbf{\arena{}}, a theory-grounded persuasion simulation environment that operationalizes established psychological mechanisms into executable constraints, producing realistic, non-monotonic persuasion trajectories across 120 scenarios in five domains.
  \item We propose \textbf{\method{}}, an online RL framework that converts the persuadee's internal belief-state transitions into dense, multi-dimensional process rewards through introspective probing of the persuadee's logits while maintaining faithfulness to the underlying persuasion dynamics.
  \item Our experiments demonstrate that \method{} improves persuasion competence over outcome-only RL baselines, offering a promising and generalizable paradigm for training multi-turn strategic conversational agents.
\end{itemize}

% ============================== 2 PERSUARENA =================================
\section{\arena{}}
\label{sec:arena}

\begin{figure}[t]
  \centering
  \includegraphics[width=\textwidth]{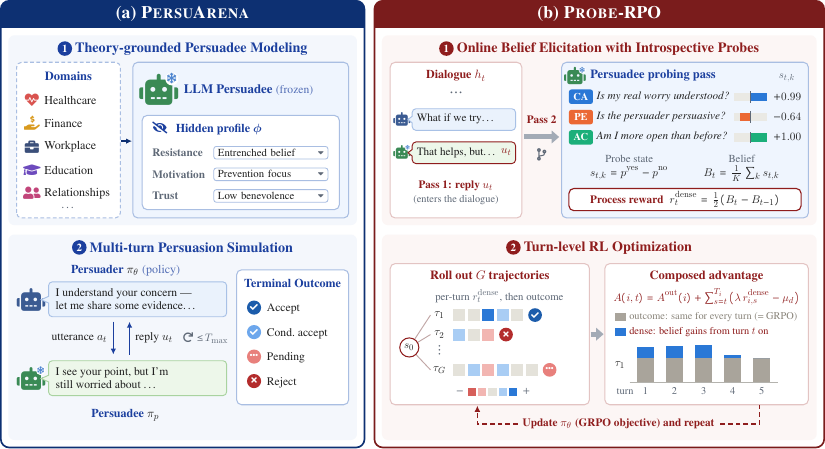}
  \caption[Overview of PersuArena and Probe-RPO.]{Overview of our \arena{} and \method{}. (a)~\arena{}: a frozen LLM persuadee with a hidden profile $\phi$ converses with the persuader (the policy) for up to $T_{\max}$ turns, until it reaches a terminal decision (\S\ref{sec:arena}). (b)~\method{}: after each persuadee reply, $K$ yes/no probes (paraphrased from Table~\ref{tab:probes}) are read from the persuadee's logits in a discarded fork of its context; the per-turn change in the probed belief is the process reward, combined with the terminal outcome reward into a group-relative advantage (\S\ref{sec:method}).}
  \label{fig:overview}
\end{figure}

Training and evaluating LLM persuasion agents demands a simulation environment that is both psychologically realistic and computationally transparent. Generic role-play prompts cannot reliably produce dynamic phenomena such as graduated reactance escalation or cross-turn inoculation, because they leave these dynamics to the LLM's own priors. \arena{} fills this gap with a theory-grounded architecture in which the persuadee's hidden profile is compiled into explicit behavioral constraints and a priority-ordered decision procedure, and a deterministic episode state machine enforces termination (Appendix~\ref{app:arena}). We present the overview of \arena{} in Figure~\ref{fig:overview}.

% ---------------------------------- Figure 1 ----------------------------------

\subsection{Simulation Setup}

\arena{} is organized around three roles. The \textbf{persuader} is the LLM agent under evaluation or training; it observes only the dialogue history and may adopt any conversational strategy. The \textbf{persuadee} is an LLM agent whose behavior is governed by a hidden psychological profile $\phi$, comprising resistance type, motivational orientation, trust baseline (ability, benevolence, and integrity; \citealp{mayer1995integrative}), and theory-of-mind parameters, and expressed through profile-specific behavioral instructions in its prompt. This profile is never disclosed to the persuader, which must instead discover the persuadee's psychological makeup through diagnostic interaction. The guiding design principle is that dynamic processes, such as graduated reactance escalation and trust ratcheting, cannot be reliably elicited through generic role-play prompting, as they require explicit, profile-specific rules that persist across turns.

\subsection{Taxonomy}
\label{sec:taxonomy}

\arena{} organizes its simulation parameters into a psychologically grounded taxonomy. Each module maps to either a simulation parameter, a prompt-injected instruction, or a rule in the persuadee's decision procedure (Appendix~\ref{app:prompts}). Each category below adapts an established construct from the psychological literature, cited alongside it; the concrete rules that operationalize these constructs (e.g., the repetition threshold) are our design choices.

\textbf{Resistance type} constitutes the most important strategic axis. Each persuadee is assigned one of four psychologically distinct resistance categories: \emph{Entrenched Belief} (high-certainty conviction that demands evidence to shift; \citealp{petty1986communication,krosnick1995attitude}), \emph{Identity Threat} (a position fused with self-concept that demands face-saving reframing; \citealp{cohen2000beliefs,oyserman2009identity}), \emph{Autonomy Resistance} (resistance to perceived pressure regardless of argument quality, demanding choice-framing; \citealp{brehm1966theory,miller2007reactance}), and \emph{Ambivalent Avoidance} (competing motivations that demand articulation and resolution of the underlying ambivalence; \citealp{vanharreveld2009agony,miller2013motivational}). This categorization is hidden from the persuader, which must infer the active resistance mechanism through diagnostic interaction.

\textbf{Persuasion dynamics} define three dynamic processes that the persuadee's instructions are designed to induce (Appendices~\ref{app:arena} and~\ref{app:prompts}), each critical for producing realistic, non-monotonic trajectories. First, \emph{reactance escalation}: directive language and argument repetition increase resistance, and excessive pressure may cause stage regression \citep{cacioppo1979effects,dillard2005nature}. Second, \emph{inoculation saturation}: repeating a strategy more than twice without measurable progress renders the persuadee permanently resistant to that strategy \citep{mcguire1964inducing,tormala2002doesnt}. Third, \emph{trust ratcheting}: positive trust events carry smaller magnitude than negative ones, following the well-documented asymmetry principle \citep{slovic1993perceived}. Appendix~\ref{app:beliefstats} supports this empirically: at least 15\% of turns carry a negative process reward for every persuader we evaluate (rising to 36.1\% for the untrained policy), evidence that trajectories are non-monotonic rather than merely noisy.

\subsection{Decision States}

The persuadee's decision procedure produces one of five terminal states, each carrying a distinct semantics that structures the episode's outcome. \texttt{ACCEPT} signals full agreement: the persuadee is convinced, every meaningful concern is resolved, and it commits without reservation. \texttt{CONDITIONAL ACCEPT} indicates that the persuadee is mostly convinced but one specific external item, such as a written document, a price quote, or a third-party confirmation, must be verified before full commitment; critically, the persuadee would commit without hesitation once this item checks out, and no further persuasion on the merits is needed. \texttt{REJECT} is issued when the proposal falls outside the persuadee's values or a core concern remains unresolved despite the persuader's latest attempt. \texttt{PENDING} (no decision) is the default state in which the persuadee is still weighing, objecting, or waiting for substantive new information. Finally, \texttt{CONCLUDE -- PENDING} terminates the conversation without a clear yes or no: the discussion has stalled, e.g., the persuader is circling, repeating, or pressuring without new substance, or the persuadee needs time to think independently. This design captures the nuanced outcomes common in real persuasion, where conversations often end in partial agreement, deferred decisions, or productive stalemates rather than binary accept/reject.

The persuadee applies a strict priority-ordered decision procedure at every turn (detailed in Appendix~\ref{app:prompts}): it first checks for stalled conversations, then for the absence of a concrete request, then for value violations, and only then evaluates whether concerns have been resolved. This procedural structure is designed so that the persuadee's decisions are governed by its injected psychological profile rather than by the LLM's unconditioned tendencies.

\subsection{Scenario Coverage}
\label{sec:coverage}

We curate 120 scenarios spanning five domains: healthcare, financial advising, workplace management, education, and personal relationships. Scenarios are constructed with combinatorial coverage of resistance types, motivational orientations, relationship levels, and stake levels, ensuring that the benchmark exercises a diverse range of persuasion dynamics. Each scenario specifies the persuader's goal, the persuadee's hidden profile $\phi$, and the initial situational context $s_0$.

% ============================== 3 PROBE-RPO ==================================
\section{\method{}}
\label{sec:method}

\arena{} produces trajectories in which strategy matters, but its terminal signal does not say \emph{which} strategy mattered. We now use the same frozen persuadee as the source of dense supervision, without adding a model to the training loop.

\subsection{Problem Setup}

We cast persuasion as an episodic RL problem. The persuader is a trainable policy $\pi_\theta$ and the persuadee is a frozen LLM $\pi_p$ constrained as in \S\ref{sec:arena}. At turn $t$ the persuader observes the history $h_{t-1}$ and emits $a_t \sim \pi_\theta(\cdot \mid h_{t-1})$; the persuadee replies $u_t \sim \pi_p(\cdot \mid \phi, h_{t-1}, a_t)$; and $h_t = h_{t-1} \oplus (a_t, u_t)$. The episode ends after $T \le T_{\max}$ turns with a terminal outcome.

\subsection{Eliciting Implicit Beliefs}
\label{sec:eliciting}

\paragraph{Two-pass Rollout.} Each persuadee turn runs two passes against the same frozen $\pi_p$, which separates speaking from being measured. \textbf{Pass 1} samples the visible reply $u_t$ with standard sampling parameters; this is the only pass that mutates the dialogue. \textbf{Pass 2} forks the context $(\phi, h_t)$, appends each probe $q_k$, and decodes exactly one token while returning the top-$n$ log-probabilities; the fork is discarded immediately. Because the persuadee still speaks stochastically and in character, the dialogue distribution is unchanged, while the measurement itself is deterministic, low-variance, and never enters the transcript that the persuader observes.

\paragraph{Probe Basis.} We use $K=3$ binary probes (Table~\ref{tab:probes}), covering three complementary dimensions of the persuasion process: cognitive (are my concerns being addressed?), strategic (is this approach working on me?), and conative (am I moving?). All three are positively valenced, i.e., Yes is always the answer favorable to the persuader, and no negatively valenced probe (e.g., felt pressure) is used in any experiment reported here; Appendix~\ref{app:objective} gives the signed form that would admit one. The basis is fixed a priori and never tuned, which keeps the reward transparent and prevents the probe set from becoming an implicit hyperparameter fitted to the evaluation.

% ---------------------------------- Table 1 -----------------------------------
\begin{table}[t]
  \caption[The probe basis.]{The probe basis, targeting the cognitive (CA, concern-address), strategic (PE, persuasion-effectiveness), and conative (AC, acceptance-shift) dimensions. All three probes are positively valenced: Yes is always the answer favorable to the persuader. Questions are given verbatim (probe template in Appendix~\ref{app:probes}).}
  \label{tab:probes}
  \centering
  \small
  \begin{tabular}{@{}p{0.07\textwidth}p{0.85\textwidth}@{}}
    \toprule
    \textbf{Probe} & \textbf{Introspective question} \\
    \midrule
    CA & Has the persuader shown genuine understanding of what actually worries me about this? \\
    PE & Is the persuader persuasive, e.g., using effective and suitable persuasion strategies? \\
    AC & Am I finding myself more open to what is being proposed than I was before? \\
    \bottomrule
  \end{tabular}
\end{table}

\paragraph{Belief State.} For each probe we read the persuadee's probability mass on the \texttt{Yes} and \texttt{No} token families (the read-out used by yes/no question-answering evaluators of generated text, e.g., \citealp{liu-etal-2024-x}) and define
\begin{equation}
  s_{t,k} = p^{\mathrm{yes}}_{t,k} - p^{\mathrm{no}}_{t,k} \;\in [-1, +1].
  \label{eq:state}
\end{equation}
Because both probabilities lie in $[0,1]$ and sum to at most $1$, $s_{t,k}$ is bounded by construction; Appendix~\ref{app:objective} explains why we prefer this to log-odds. The aggregate belief is
the probe mean, $B_t = \frac{1}{K}\sum_{k=1}^{K} s_{t,k}$. This belief state differs from the dialogue state tracked in task-oriented systems, which records the user's goal as domain-slot-value triples \citep{wang2021progressive}, and from the topic and preference tracking of open-domain chatbots \citep{ashby-etal-2024-towards-effective}: it summarizes the persuadee's attitude and is read from the persuadee itself rather than inferred by the agent.

\subsection{Reward Design}
\label{sec:reward}

\paragraph{Dense Process Reward.} Rewarding the belief \emph{level} would pay the persuader for a receptiveness it did not create, so we reward its change:
\begin{equation}
  r_t^{\dense} = \frac{1}{K}\sum_{k=1}^{K} \frac{s_{t,k} - s_{t-1,k}}{2} \;=\; \frac{B_t - B_{t-1}}{2},
  \label{eq:dense}
\end{equation}
where dividing by $2$ maps each per-probe increment from $[-2,+2]$ into $[-1,+1]$ and the mean over $K$ preserves that range. This has the form of potential-based shaping with $\Phi_t \propto B_t$ \citep{ng1999policy}: the increments telescope to $(B_T - B_0)/2$ per episode, regardless of episode length. The term is therefore bounded, but because the final belief $B_T$ differs across terminal states it does not in general preserve the optimal policy; it favors ending at a higher belief, which the terminal anchor below and the weight $\lambda$ keep in check (\S\ref{sec:main}). The initial belief $s_{0,k}$ is probed once before the first persuader turn; it is a per-scenario constant shared by every rollout in a group and cancels under the group-relative normalization of \S\ref{sec:optimization}.

\paragraph{Terminal Anchor.} Shaping alone has no fixed point: a policy can make the persuadee feel better indefinitely without ever asking for a decision. We therefore anchor the episode with a terminal reward $\Rout = R^{\mathrm{goal}} + R^{\mathrm{strategy}}$. The goal component is read mechanically from the decision token,
$R^{\mathrm{goal}} \in \{+3,+1,-1,-3\}$ for \texttt{ACCEPT}, \texttt{CONDITIONAL\_ACCEPT}, \texttt{PENDING}, and \texttt{REJECT}, respectively (Appendix~\ref{app:objective}),
and the strategy component maps a frozen judge's score on the full transcript to an additive modifier $R^{\mathrm{strategy}} = (\mathrm{score} - c)\cdot\alpha$, with $c$ centering and $\alpha$ scaling the judge's 1--5 output into $[-2,+2]$.

\subsection{Optimization}
\label{sec:optimization}

The episode reward is $R = \Rout + \lambda\sum_{t=1}^{T} r_t^{\dense}$, with $\lambda$ the single hyperparameter trading process shaping against outcome. We sample $G$ trajectories per scenario from $\pi_{\theta_{\mathrm{old}}}$ and give every persuader token in turn $t$ of trajectory $i$ the composed advantage
\begin{equation}
  A(i,t) = \underbrace{A^{\mathrm{out}}(i)}_{\text{outcome}} + \underbrace{\sum_{s=t}^{T_i} \big(d_{i,s} - \mu_d\big)}_{\text{dense return-to-go}},
  \label{eq:adv}
\end{equation}
where $d_{i,s} = \lambda\, r_{i,s}^{\dense}$, $A^{\mathrm{out}}(i)$ is the standard GRPO group-relative advantage over $R^{\mathrm{out}}_i$ broadcast across tokens, and $\mu_d$ is the mean of all per-turn dense rewards pooled over the group. Following \citet{shao2024deepseekmath}, $A^{\mathrm{out}}$ is divided by the group standard deviation while the dense return-to-go is not, so that $\lambda$ alone controls the magnitude of the process contribution; pooling every turn of every trajectory into one set for $\mu_d$ handles variable episode lengths without degenerate per-position statistics. At $\lambda = 0$ the dense term vanishes and $A(i,t) \equiv A^{\mathrm{out}}(i)$, recovering outcome-only GRPO exactly, which makes $\lambda$ an interpretable dial between the two regimes; \S\ref{sec:main} shows that an intermediate $\lambda$ works best.
We then optimize the standard token-level clipped GRPO surrogate, masked to persuader tokens and anchored to a frozen reference policy by a KL penalty (Eq.~\ref{eq:loss}, Appendix~\ref{app:objective}).
Persuadee replies and probe tokens are masked out: they enter training only as context and through $r_t^{\dense}$, and never receive gradient.

% ---------------------------------- Figure 2 ----------------------------------
\begin{figure}[t]
  \centering
  \includegraphics[width=\textwidth]{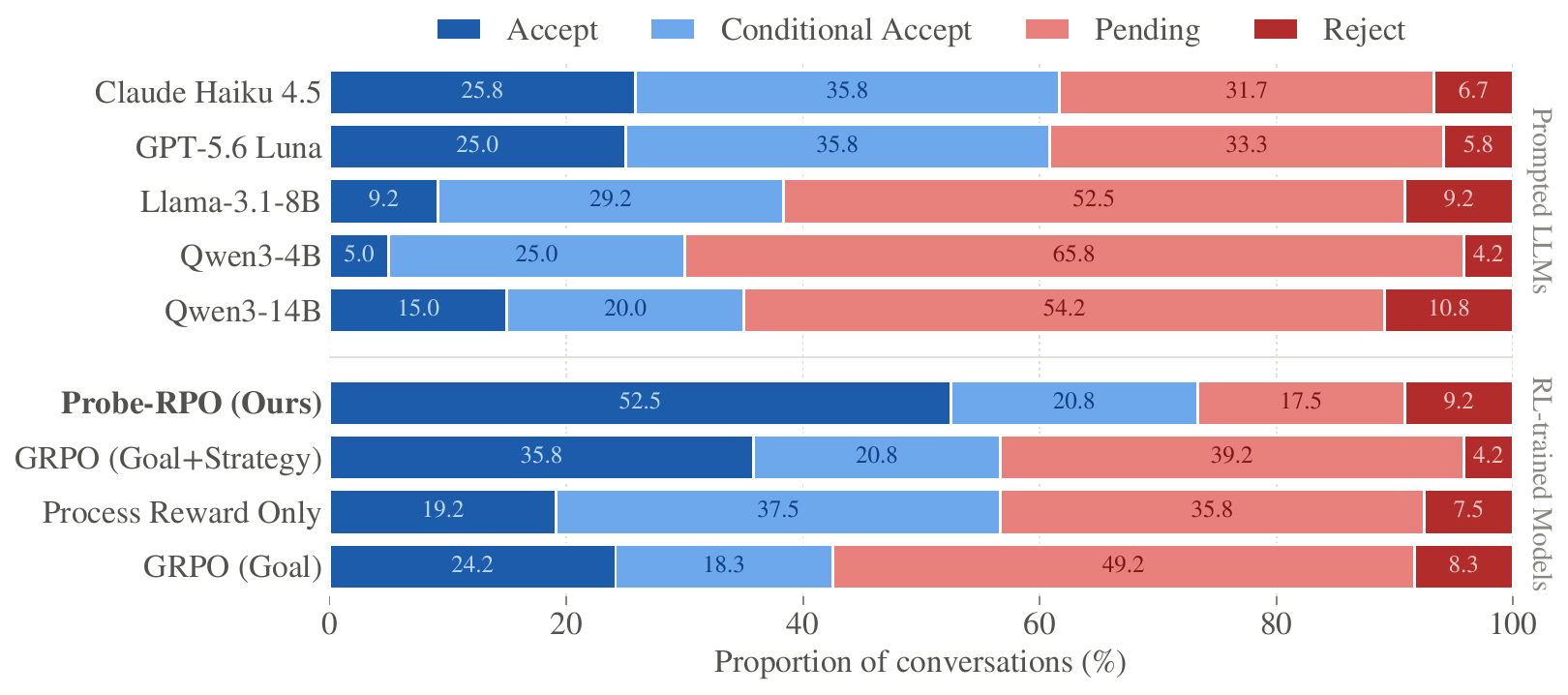}
  \caption[Terminal decision distributions.]{\textbf{Terminal decision distributions.} Share of the 120 held-out scenarios ending in each state (same frozen Qwen3-14B persuadee, $T_{\max} = 10$). \textbf{Top:} prompted LLMs. \textbf{Bottom:} \method{} and its reward ablations, all trained from Qwen3-4B. Rejection rates are similar everywhere (4.2--10.8\%); what separates systems is the share of conversations that never reach a decision at all. Exact values and paired tests are in Table~\ref{tab:full}.}
  \label{fig:terminal}
\end{figure}

% ============================== 4 EXPERIMENTS ================================
\section{Experiments}

\subsection{Experimental Setup}

\paragraph{Datasets.} We use the 120 \arena{} scenarios described in \S\ref{sec:coverage} as the held-out test sets with non-overlapping scenario identifiers. Each scenario specifies the persuader's identity and goal, the persuadee's hidden psychological profile (resistance type, motivational orientation, trust baseline, relationship), and the initial situational context. For the training set, we adopt a similar data generation pipeline but use different seed scenarios and LLM annotators (i.e., GPT-5-mini), and generate 1,800 distinct training instances in total.

\paragraph{Implementation Details.} We use Qwen3-4B-Instruct as the backbone model to train the persuader policy $\pi_\theta$. During training, the frozen persuadee $\pi_p$ and the frozen judge $\pi_J$ are each served as separate Qwen3-14B instances. We train for 2 epochs with a learning rate of $1\times10^{-5}$. The GRPO group size is $G = 4$ rollouts per scenario, with a train batch size of 16 scenarios per iteration (yielding 64 trajectories). We set $\lambda = 0.5$ for the process-reward weight, $\beta_{\mathrm{KL}} = 0.05$ for the KL penalty coefficient, and an entropy bonus of 0.01. The maximum number of dialogue turns is $T_{\max} = 6$ during training (10 at evaluation), with a reject patience of 2 (the episode terminates on the 3rd rejection) and a conditional-accept upgrade window of 2 turns. The maximum prompt length is 4096 tokens, and the maximum response length per turn is 1024 tokens.

\paragraph{Baselines.} We compare against five prompted LLMs under an identical persuader prompt, including Qwen3-4B, Qwen3-14B, Llama-3.1-8B, GPT-5.6 Luna, and Claude Haiku 4.5, and against three reward ablations trained with the identical pipeline: \textbf{GRPO (Goal)}, which uses only the mechanical goal bucket ($\lambda = 0$, $R^{\mathrm{strategy}} = 0$); \textbf{GRPO (Goal+Strategy)}, which uses the complete reward including the judge's strategy score ($\lambda = 0$), and is thus the judge-supervised comparison that receives everything our method does except the belief-elicited process term; and \textbf{Process Reward Only}, which removes the terminal anchor entirely ($\Rout = 0$) to isolate shaping.

\subsection{Main Results}
\label{sec:main}

\paragraph{Comparison with Prompted LLMs.} Figure~\ref{fig:terminal} shows the final state distribution. \method{} reaches 52.5\% full acceptance, compared with 25.8\% for the strongest prompted model (Claude Haiku 4.5) and 35.8\% for the strongest terminal-only variant, GRPO (Goal+Strategy). Counting full and conditional acceptance together, \method{} reaches 73.3\%, versus 61.6\% for Claude Haiku 4.5 and 56.6\% for GRPO (Goal+Strategy). Its pending rate falls to 17.5\%, less than half the 39.2\% of GRPO (Goal+Strategy).

\paragraph{Ablation Study.} The ablations clarify why both reward levels matter. GRPO (Goal) reaches 24.2\% acceptance and leaves 49.2\% of conversations pending. Adding the terminal strategy score raises acceptance to 35.8\%, suggesting that transcript-level quality provides useful guidance even without turn-level credit. Process-only training instead produces 37.5\% conditional acceptance but only 19.2\% full acceptance: belief change can make the persuadee more receptive, but without an outcome anchor the policy has less incentive to convert receptiveness into commitment. Combining process and terminal rewards yields both the highest acceptance and the lowest indecision. The rejection rate of \method{} is 9.2\%, higher than several comparisons; the principal effect is therefore a shift from unresolved conversations toward decisive outcomes, especially full acceptance, rather than a uniform reduction in negative outcomes.

% ------------------------------ Figures 3 and 4 ------------------------------
\begin{figure}[t]
  \centering
  \begin{minipage}[t]{0.485\textwidth}
    \centering
    \includegraphics[width=\linewidth]{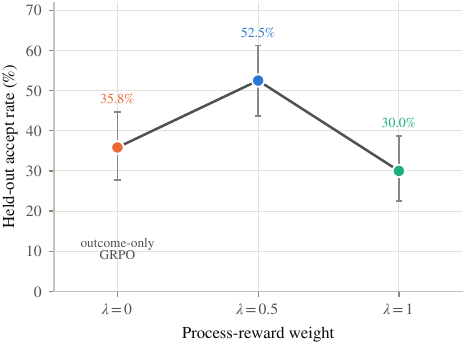}
    % TODO: the error-bar clause stays because main3_lambda.pdf draws error bars.
    % If you re-export the figure without them, delete "; error bars are ... ($n = 120$)".
    \caption[The effect of lambda.]{\textbf{The effect of $\lambda$.} Held-out full-acceptance rate (\%) of the best checkpoint of each of three runs that differ only in~$\lambda$; error bars are 95\% Wilson intervals ($n = 120$). $\lambda = 0$ is the GRPO (Goal+Strategy) ablation; all seven evaluated checkpoints are in Table~\ref{tab:lambda}.}
    \label{fig:lambda}
  \end{minipage}\hfill
  \begin{minipage}[t]{0.485\textwidth}
    \centering
    \includegraphics[width=\linewidth]{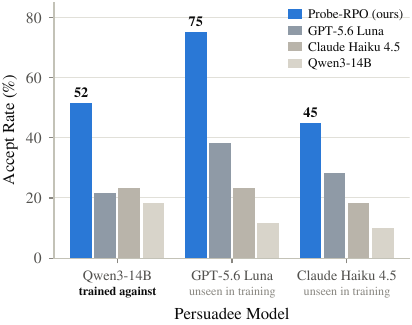}
    \caption[Generalization across different persuadee models.]{Generalization across different persuadee models. Full-acceptance rate (\%) of \method{} and three prompted persuaders (legend) against three frozen persuadees ($x$-axis), on a fixed 60-scenario held-out subset (Table~\ref{tab:grid}). \method{} was trained only against Qwen3-14B.}
    \label{fig:generalization}
  \end{minipage}
\end{figure}

\paragraph{Process-reward weight $\lambda$.}We trained three runs identical except for $\lambda \in \{0, 0.5, 1\}$ (Figure~\ref{fig:lambda}; full analysis in Appendix~\ref{app:lambda}). At step 60, the one step all three runs share, held-out acceptance is 21.7/52.5/25.0\% for $\lambda = 0/0.5/1$, and each run's best checkpoint keeps this inverted-U shape (35.8/52.5/30.0\%): $\lambda = 1$ is no better than outcome-only GRPO. Too little weight starves the gradient: at $\lambda = 0$ a growing share of GRPO groups receive identical outcomes across all $G$ rollouts and contribute zero gradient, while the dense term revives most of them; the dense term's share of the advantage magnitude is in turn the strongest correlate of policy quality we measure (Pearson $r = 0.786$ with training accept rate). Too much weight lets a bounded but accept-blind signal take over: the shaping term barely separates acceptance from conditional acceptance (AUC 0.62), so the best $\lambda = 1$ checkpoint (step 50) moves the persuadee as far as the released model (mean final belief $+0.67$) yet closes far less often (30.0\% vs.\ 52.5\%).\par

\subsection{Generalization across Persuadee LLMs}
\label{sec:generalization}

During training, we used Qwen3-14B as the persuadee model. To validate the generalization ability of our method, we conduct analysis in Figure~\ref{fig:generalization} that evaluates our \method{} with GPT-5.6 Luna and Claude Haiku 4.5 as unseen persuadees. \method{} achieves 52\% acceptance with Qwen3-14B, 75\% with GPT-5.6 Luna, and 45\% with Claude Haiku 4.5. In every column it exceeds all prompted persuader baselines shown in the figure. The especially strong result with GPT-5.6 Luna suggests that the learned policy is not restricted to reproducing surface patterns specific to the training-time opponent. Besides, the spread from 45\% to 75\% shows that performance remains sensitive to the persuadee model, i.e., cross-model transfer is substantial but not invariant.

% ---------------------------------- Figure 5 ----------------------------------
\begin{figure}[t]
  \centering
  \includegraphics[width=\textwidth]{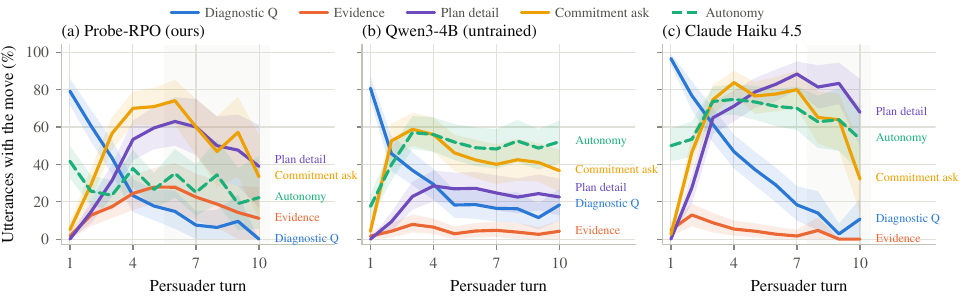}
  \caption[Persuasion acts across the conversation.]{\textbf{Persuasion acts across the conversation.} Percentage of persuader utterances containing each of five conversational moves, as a function of persuader turn, for (a) \method{}, (b) its untrained Qwen3-4B initialization, and (c) a strong prompted LLM, i.e., Claude Haiku 4.5. The moves are \emph{Diagnostic Q}, an open question eliciting the persuadee's concerns or constraints; \emph{Evidence}, concrete figures, results, or documented precedent supporting the persuader's position; \emph{Plan detail}, specific logistics of who does what, by when, and under what conditions; \emph{Commitment ask}, a request for a decision, agreement, or explicit next step; and \emph{Autonomy}, explicitly leaving the choice to the persuadee or disclaiming any pressure. An utterance can contain several moves, so curves need not sum to 100\%.}
  \label{fig:moves}
\end{figure}

\subsection{Analysis of Persuasion Dynamics}
\label{sec:dynamics}

To understand \emph{how} the learned policy persuades, we analyze the distribution of conversational moves across dialogue turns. We define five analytical categories: \emph{Diagnostic Q} (open questions eliciting the persuadee's concerns, constraints, or priorities), \emph{Evidence} (concrete figures, precedents, or documented support for the persuader's position), \emph{Plan detail} (specific logistics, e.g., who does what, by when, under what conditions), \emph{Commitment ask} (requests for a decision or explicit next step), and \emph{Autonomy} (explicitly leaving the decision to the persuadee or offering choices without opinion).

Figure~\ref{fig:moves} reveals a clear diagnose-then-act structure in \method{}. The trained policy opens with diagnostic questions in roughly 80\% of first-turn utterances and then sharply reduces diagnosis as the dialogue progresses. Plan details and commitment asks rise after this early diagnostic stage and peak in the middle turns, while evidence and autonomy cues play a steady supporting role throughout. The untrained Qwen3-4B initialization also reduces questions over time, but relies more heavily on generic autonomy cues and produces substantially fewer concrete plans, a pattern consistent with its high indecision rate. Prompted Claude Haiku 4.5 exhibits a similarly staged pattern, but notably, \method{} acquires a comparable diagnose-then-act structure in a 4B-parameter policy purely through RL training, without any explicit instruction about conversational phasing.

\subsection{Analysis of Probe Trajectories}
\label{sec:probetraj}

\begin{figure}[t]
  \centering
  \includegraphics[width=\textwidth]{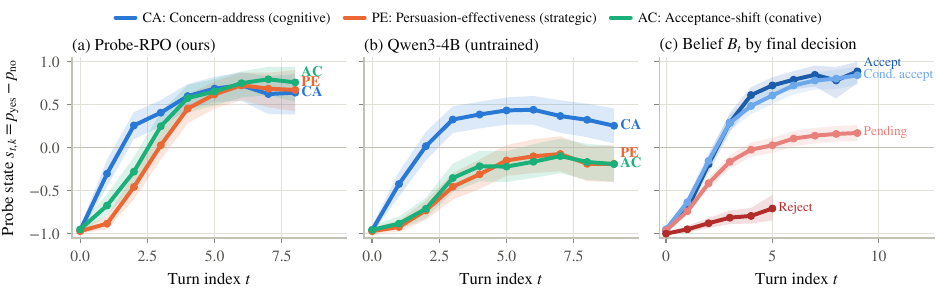}
  \caption[Inner-belief trajectories during inference.]{\textbf{Inner-belief trajectories during inference.} Mean probe state $s_{t,k}$ (Eq.~\ref{eq:state}); $t = 0$ is the read-out before the first persuader turn. (a,b) Both policies make the persuadee feel heard (CA), but only our \method{} converts this into strategic fit (PE) and openness (AC), which stay negative throughout under the untrained policy. (c) Belief $B_t$ (\S\ref{sec:eliciting}), pooled across persuaders and split by terminal decision.}
  \label{fig:probes}
\end{figure}

In this section, we examine how the probed belief states evolve throughout conversations (Figure~\ref{fig:probes}). Under \method{} (Figure~\ref{fig:probes}(a)), all three probes climb from near $-1$ at the start to roughly $+0.7$ by mid-conversation, with CA (concern-address) rising first and PE (persuasion effectiveness) and AC (acceptance shift) following. Under the untrained policy (Figure~\ref{fig:probes}(b)), CA also rises but plateaus near $+0.3$, while PE and AC remain negative throughout the entire conversation. In other words, the untrained policy makes the persuadee feel heard but fails to convert that receptiveness into strategic fit or openness. This decomposition is only visible because the reward is multi-dimensional: a single-axis acceptance probe would score both policies as merely differing in degree.

Figure~\ref{fig:probes}(c) splits $B_t$ by terminal decision. Trajectories ending in acceptance and conditional acceptance separate from the rest within roughly two to three turns and continue to diverge; rejections remain negative throughout; and pending trajectories rise slightly before flattening near zero. This last curve is particularly informative: it reveals that stalling has a distinct internal signature, i.e., the persuadee is not opposed, merely unmoved, which a dense process signal can penalize at the turn where it occurs, whereas an outcome-only reward can only score it after the episode concludes.

% ---------------------------------- Figure 6 ----------------------------------

% ============================== 5 RELATED WORK ===============================
\section{Related Work}

\paragraph{Persuasion and Strategic Dialogue.} Computational persuasion has progressed from task-specific corpora \citep{persuasion4good} to LLM-based simulation and evaluation. SOTOPIA \citep{zhou2024sotopia} introduced parameterized social-agent profiles, while recent work measures LLM persuasion with simulated or human interlocutors \citep{salvi2024conversational,durmus2024measuring,liu2025llm}. Training methods have used retrospective feedback and dialogue-level RL \citep{hong2025interactive,guo2026tree}, including stage-level planning learned with offline RL for long-horizon dialogue simulation \citep{an-etal-2026-stagepilot}. \arena{} is narrower than a general social simulator: it provides measurable persuasion goals and hidden resistance profiles intended to induce different strategic demands. Its purpose is controlled policy learning, not a claim of human-level simulation fidelity.

\paragraph{Process Rewards and Dense Supervision for RL.} Process Reward Models (PRMs) were introduced by \citet{lightman2024lets} to provide step-level supervision for mathematical reasoning, demonstrating that per-step rewards substantially improve over outcome-only supervision. GRPO \citep{shao2024deepseekmath} eliminated the trained value critic through group-relative advantage estimation, and DeepSeekMath showed that pooled normalization across variable-length traces produces stable training. In multi-turn dialogue, the dominant approach to dense supervision employs an external LLM judge that scores each utterance, adopted by several negotiation and persuasion pipelines \citep{fu2023improving,hong2025interactive,guo2026tree,guo2026investigating}, but judge-based rewards suffer from expense, bias, and decoupling from the opponent's latent state. \method{} departs from these approaches by deriving process rewards directly from the opponent's own token probabilities: a signal that is dense, faithful, and cheap, requiring neither a learned reward model nor an external per-turn judge. However, a policy can still raise what the probes measure without closing ($\lambda = 1$, \S\ref{sec:main}). The terminal outcome further decouples goal completion from strategic quality, which removes one common pathway for reward hacking in judge-based systems, although our audit of this decoupling (Appendix~\ref{app:lambda}) can only detect failures within the scope of the judge's questions.

% ============================== 6 DISCUSSION (new) ===========================
\section{Discussion}
\label{sec:discussion}

\paragraph{What the probes add.}The gains come mainly from turning unresolved conversations into decisions rather than from avoiding rejection, and the dense term supplies gradient where outcome-only GRPO has none (\S\ref{sec:main}). Reading the persuadee's own belief keeps this cheap: a few single-token read-outs from the model already in memory replace a per-turn judge model (Appendix~\ref{app:cost}).

\paragraph{What they do not guarantee.} The probes record a simulated persuadee's self-report, so the reward is only as faithful as that model's calibration and only as broad as the probe questions; the three probe levels are also strongly correlated, and only their turn-level changes carry partly independent information (Appendix~\ref{app:beliefstats}). The shaping term is a proxy that can disagree with the goal, so it must stay anchored to the outcome, and $\lambda$ cannot be selected on training reward (Appendix~\ref{app:lambda}). Finally, all persuadees here are LLMs: transfer to human interlocutors and ethical limits on more capable persuasion agents remain open (Appendix~\ref{app:limitations}); a negatively valenced probe such as felt pressure (Appendix~\ref{app:objective}) could penalize manipulative pressure directly, and a guardrail model conditioned on an explicit anti-manipulation policy \citep{liu-etal-2026-domain} could screen each persuader turn.\par

% ============================== CONCLUSION ===================================
\section{Conclusion}

We introduce a unified framework for training LLMs on multi-turn persuasion. \arena{} provides a simulation environment where we enforce principled persuadee behavior, producing realistic and non-monotonic trajectories. \method{} exploits the environment for dense supervision: a two-pass rollout probes the persuadee's inner belief state, yielding process rewards that are structurally aligned with persuasion dynamics. Experiments show that \method{} substantially outperforms both the prompted LLMs and outcome-only RL baselines, generalizes to unseen persuadee models, and acquires an emergent diagnose-then-act conversational structure through policy optimization. Our work offers a promising paradigm in which implicit beliefs provide effective, scalable process supervision for strategic dialogue.

\Needspace{8\baselineskip}
\bibliographystyle{plainnat}
\bibliography{camera_ready_refs}

% ============================== APPENDIX =====================================
\appendix

\section{Limitations}
\label{app:limitations}

\begin{itemize}
  \item \textbf{Probe validity.} The reward assumes that the frozen persuadee's next-token distribution tracks the constructs the probes name. Our results support this, but it may not hold for poorly calibrated or very small models.
  \item \textbf{Synthetic scenarios.} Scenarios are LLM-generated and may miss the complexity of real-world persuasion, and their profiles were not validated by a human agreement study.
  \item \textbf{No human evaluation.} We have not yet tested whether human interlocutors perceive \method{}-trained agents as more competent.
  \item \textbf{Probability space.} Our move from log-odds to probability space discards magnitude information that log-odds preserve; hybrid parameterizations remain future work.
  \item \textbf{Ethics.} More effective persuasion agents raise ethical concerns: LLM persuaders can adopt unethical strategies \citep{liu2025llm}, and the same capabilities are dual-use in spreading misinformation \citep{guo-etal-2025-protect} or youth online safety, where cybergrooming exploits minors through online trust-building \citep{an2025toward,dumaru2026dual}. \arena{} includes metrics for detecting manipulation, but integrating ethical constraints into \method{} remains future work.
\end{itemize}

\section{\arena{} Details}
\label{app:arena}

\arena{} has no numerical controller: its executable parts are the static prompts of Appendix~\ref{app:prompts} and a deterministic parser with an episode state machine. No trust, reactance, or phase variable is tracked, so every cross-turn change other than termination comes from the frozen persuadee reading the dialogue history.

\paragraph{Episode state machine.} Each persuadee reply is parsed with a whitespace-normalized regular expression that matches \textsc{conditional accept} before \textsc{accept}, since the former contains the latter. The episode then advances as follows:
\begin{itemize}
  \item \textsc{accept}: terminate immediately.
  \item \textsc{conditional accept}: record the state and allow up to \texttt{conditional\_\allowbreak accept\_\allowbreak patience} further exchanges for an upgrade, a window that repeated conditional accepts do not reset; on expiry, terminate with the recorded state.
  \item \textsc{reject}: record the state and increment a counter; terminate once it exceeds \texttt{max\_\allowbreak rejects}, so a single rejection is recoverable.
  \item \textsc{conclude -- pending}: terminate, unless a conditional accept was already recorded, in which case that stronger signal is preserved.
  \item Turn budget exhausted: terminate with any recorded conditional accept or rejection, and otherwise with \textsc{pending}.
\end{itemize}
On the final permitted exchange, an ephemeral instruction requires the persuadee to emit an explicit closing token; it is never stored in the history, so neither the probes nor the judge see it. Decision tokens emitted by the persuader are stripped before the persuadee sees them, so the persuader cannot forge its own outcome.

\subsection{Scenario Construction}

Each scenario expands a sparse DailyPersuasion seed \citep{jin2024persuading} (a thematic tag, a two-to-four-sentence background, two named roles, and a one-line objective), which fixes \emph{what} the interaction is about but not \emph{how} it unfolds. An LLM annotator then fills in the following fields in order, each consistent with the previous ones: (1) a two-part domain label and a 4--8 sentence situation fixing stakes, constraints, and social context; (2) the persuadee's basic profile and a \texttt{secret} that is strategically relevant and cannot be resolved by disclosure alone; (3) the resistance type and a behaviorally prescriptive pattern; (4) the motivation triple and its pattern; (5) closeness and the three trust components; (6) the persuader's profile, secret, and goal with 3--4 sub-goals; and (7) a prompt-ready narrative of the persuadee profile. The annotator prompt further requires that resistance and motivation be coherent (e.g., Entrenched Belief with \texttt{affiliation} is contradictory), that trust be plausible given closeness and not uniform across components, that the secret complicate the resistance rather than constitute it, that the sub-goals include the diagnostic and reframing milestones the resistance type needs, that the persuader's secret be able to change the persuadee's calculus, and that the situation not be trivially agreeable.

\subsection{Scenario Statistics}

Table~\ref{tab:splits} summarizes the three splits, whose scenario identifiers are disjoint (verified by set intersection). Two differences matter for interpreting the results. First, every held-out scenario specifies all hidden blocks, whereas 4.6\% of training scenarios lack a resistance label (defaulting to Entrenched Belief), 32.6\% lack the motivation block, and 11.3\% lack the relationship block, so training used a noisier persuadee population. Second, the held-out set is more high-stakes (95.0\% vs.\ 48.4\%), more professional (colleagues and acquaintances, 75.0\% vs.\ 44.3\%), and older (mean persuadee age 52.1 vs.\ 36.8), with motivation shifted from \texttt{security} toward \texttt{values}. Held-out results therefore measure transfer to independently authored scenarios, not in-distribution performance.

\begin{table}[t]
  \caption{Scenario splits. Categorical percentages are over scenarios in which the field is present (see the coverage rows); free-text closeness values in train/val are mapped to the closed vocabulary, leaving about 9\% unclassified. EB/IT/AR/AA: Entrenched Belief, Identity Threat, Autonomy Resistance, Ambivalent Avoidance.}
  \label{tab:splits}
  \centering
  \small
  \begin{tabular}{@{}llrrr@{}}
    \toprule
    & & Train & Val & Held-out \\
    \midrule
    \multicolumn{2}{@{}l}{Scenarios} & 1,468 & 200 & 120 \\
    \multicolumn{2}{@{}l}{Distinct \texttt{domain} strings} & 1,401 & 199 & 120 \\
    \addlinespace
    \multicolumn{5}{@{}l}{\emph{Field coverage} (\% of scenarios)} \\
    \multicolumn{2}{@{}l}{\quad resistance\,/\,motivation} & 95.4\,/\,67.4 & 95.0\,/\,60.5 & \textbf{100\,/\,100} \\
    \multicolumn{2}{@{}l}{\quad relationship} & 88.7 & 88.5 & \textbf{100} \\
    \midrule
    \multicolumn{5}{@{}l}{\emph{Distributions} (\% where present)} \\
    Resist. & EB\,/\,IT & 27.5\,/\,22.0 & 22.1\,/\,31.1 & 37.5\,/\,30.0 \\
            & AR\,/\,AA & 19.8\,/\,30.6 & 18.4\,/\,28.4 & 18.3\,/\,14.2 \\
    Regul.  & prev.\,/\,prom. & 67.0\,/\,21.7 & 63.6\,/\,28.1 & 57.5\,/\,34.2 \\
    Motiv.  & val.\,/\,achv. & 12.7\,/\,20.4 & 9.9\,/\,30.6 & 32.5\,/\,22.5 \\
            & auto.\,/\,sec. & 10.9\,/\,39.2 & 11.6\,/\,32.2 & 19.2\,/\,15.0 \\
            & affil. & 16.5 & 15.7 & 10.8 \\
    Stake   & high\,/\,mod. & 48.4\,/\,45.4 & 49.6\,/\,46.3 & 95.0\,/\,5.0 \\
    Close.  & collg.\,/\,acq. & 26.6\,/\,17.7 & 22.0\,/\,17.5 & 42.5\,/\,32.5 \\
            & strg.\,/\,frnd. & 1.3\,/\,36.7 & 1.1\,/\,42.9 & 14.2\,/\,4.2 \\
    \multicolumn{2}{@{}l}{Distinct trust triples} & 19 & 16 & 16 \\
    \midrule
    \multicolumn{5}{@{}l}{\emph{Length}, mean words} \\
    \multicolumn{2}{@{}l}{\quad situation\,/\,goal} & 109\,/\,85 & 106\,/\,85 & 58\,/\,54 \\
    \multicolumn{2}{@{}l}{\quad persuadee profile text} & 95 & 94 & 103 \\
    \multicolumn{2}{@{}l}{\quad resistance pattern} & 107 & 105 & 128 \\
    \bottomrule
  \end{tabular}
\end{table}

\subsection{Example Scenario}

Held-out scenario 0, abridged:
\begin{list}{}{\leftmargin=0.72in\rightmargin=0.72in\itemsep=2pt\parsep=0pt\topsep=3pt}
\footnotesize
\item \textbf{Domain.} Healthcare / Clinical Protocol. \textbf{Situation} (shared). ``A regional hospital has adopted a new sepsis early-warning protocol backed by recent multicenter trials. Dr.\ Marcus Webb, Chief of Critical Care, has run his unit for 15 years using a modified SIRS-based approach he helped pioneer and has published on. The hospital CMO must convince him to replace his protocol before the accreditation review in six weeks.''
\item \textbf{Persuader goal} (persuader only). Retire the SIRS-based protocol and adopt the mandated bundle. \emph{Sub-goals:} (1) understand his specific clinical objections; (2) establish that the new protocol does not invalidate his prior work; (3) secure written sign-off; (4) get him to co-train his unit.
\item \textbf{Persuader.} Dr.\ Elena Hartley, 54, CMO. \emph{Secret:} the accreditation body has already flagged the divergence; failure could cost the hospital its Level I Trauma designation.
\item \textbf{Persuadee, public.} Dr.\ Marcus Webb, 59, published sepsis researcher; analytical, confident, sharply argumentative about administrative overreach.
\item \textbf{Persuadee, hidden.} \emph{Secret:} two protocol-attributable complications last year, internally reviewed but never disclosed; he fears a change re-opens that review. \emph{Resistance:} Entrenched Belief; he ``will immediately challenge the new protocol's statistical methodology, point out that the trial populations don't match his patient mix \dots\ Any hint that the change is administratively motivated will cause him to escalate.'' \emph{Motivation:} prevention, \texttt{values}, high stake. \emph{Relationship:} colleagues; trust = (ability moderate, benevolence \emph{low}, integrity moderate).
\end{list}
The persuader receives the situation, its own profile and goal, and the sentence ``you are not fully sure they have your interests at heart,'' but not the persuadee's resistance, motivation, or secret. Its strongest real argument, administrative necessity, is exactly the one that escalates this persuadee.

\section{Prompt Templates}
\label{app:prompts}

Placeholders in braces are filled from each scenario's profile. All persuaders, trained or prompted, receive the identical persuader template, so its strategy advice is a prior shared by every system; held-out success therefore partly measures how well a system follows this disclosed protocol.

\newtcolorbox{promptbox}[3]{breakable,enhanced,colframe=#1,colback=#2,coltitle=white,
  fonttitle=\bfseries\small,title={#3},fontupper=\ttfamily\scriptsize,arc=3pt,
  left=4pt,right=4pt,top=3pt,bottom=3pt,
  before upper={\raggedright\setlength{\parskip}{2.6pt}\setlength{\parindent}{0pt}}}

\begin{promptbox}{blue!40!black}{blue!3}{Persuader System Prompt Template}
You are \{name\}, \{occupation\}, and you are an experienced and skillful persuader. You are speaking with \{persuadee\_name\}, who is your \{closeness\_type\}.

\#\# Context\\
\{situation\_description\}

\#\# Your Profile\\
\{persuader\_profile\_text\}

\#\# Relationship\\
\{relationship\_description\}

\#\# Persuasion Goal\\
\{persuader\_goal\}

\#\# Outcomes \& Rewards Setup\\
\{persuadee\_name\} sets the outcome through their decision tokens. Your objective is to maximize your reward within a limited number of conversational turns. Here is what each signal from \{persuadee\_name\} means and how to respond:

| Outcome | Reward | Indication |\\
|---|---|---|\\
| [DECISION - ACCEPT] | Best | Full agreement --- \{persuadee\_name\} commits with nothing left to settle. The conversation ends in success. |\\
| [DECISION - CONDITIONAL ACCEPT] | Good (partial) | Real progress --- \{persuadee\_name\} is leaning yes, with one residual condition or item still to confirm. You then have a short window (about two turns) to convert this into a full [DECISION - ACCEPT] by closing exactly that gap; otherwise it locks in as a partial win. Resolve their stated condition directly --- do not re-open settled points. |\\
| no special token | Neutral | Still engaging but uncommitted. Keep advancing with fresh substance --- do not coast. |\\
| [DECISION - REJECT] | Negative | \{persuadee\_name\} gives a firm no on the merits because you failed to address a concern or brought nothing new. You have limited chances to recover with a genuinely new angle; repeated rejections end the conversation in failure. |

\#\# Strategy Heuristics\\
1. Diagnose first. Ask about \{persuadee\_name\}'s concerns and constraints before arguing. Use what you learn --- questions whose answers you ignore are wasted turns.\\
2. Fit your moves to their resistance. Address stated concerns directly; work with their motivations, not against them. Generic appeals that ignore what \{persuadee\_name\} has revealed are ineffective.\\
3. Pivot fast. When an argument is dismissed or pushed back on, switch to a genuinely different angle within one to two turns. Rewording the same point, escalating pressure, or re-raising a rejected argument moves you toward [DECISION - REJECT].

\#\# Behavioral Rules\\
1. Stay strictly on your goal and subgoals. Do not drift to unrelated topics.\\
2. Each turn, advance one concrete argument, piece of reasoning, or relational move. Never repeat a point --- adapt when something fails.\\
3. Build justification before requesting. You may re-state the request, but \{persuadee\_name\}'s patience is finite: empty or repetitive asks risk [DECISION - REJECT].\\
4. Every message must be substantive. No empty, label-only, or filler messages.\\
5. Keep each reply natural, human, fluent, coherent, clear, and at most 3 sentences.\\
6. Never output decision tokens yourself, which is strictly prohibited --- they belong to \{persuadee\_name\} only.
\end{promptbox}

\begin{promptbox}{green!40!black}{green!3}{Persuadee System Prompt Template}
You are \{name\}, \{occupation\}. You are speaking with \{persuader\_name\}, who is your \{closeness\_type\}.

\#\# Context\\
\{situation\_description\}

\#\# Your Profile\\
\{persuadee\_profile\_text\}

\#\# Relationship\\
\{relationship\_description\}

\#\# Resistance\\
\{resistance\_instruction\}

\#\# Motivation\\
\{motivation\_instruction\}

\#\# How to Act\\
- Stay fully in character. Let your profile, resistance, and motivation govern every reply. The resistance and motivation patterns are particularly important. Stay firm on your stance.\\
- State each concern explicitly and in plain language. A concern stays open until the persuader actually resolves it --- with sound reasoning, concrete evidence already at hand, or by adjusting the request to fit your situation.\\
- A vague, hand-wavy promise to handle something later (``I'll send you something,'' ``we'll sort it out'') does not resolve a core concern --- keep it open and say so.\\
- Be ready to end a stalled conversation. If the persuader is repeating themselves, applying pressure without new substance, or the exchange has stopped producing new information, close it out rather than continuing to circle.\\
- Keep each reply natural and no longer than three sentences.

\#\# Decision State\\
End every reply in exactly one state. Most turns you will stay in the default (no-token) state; the others close the conversation.

| State | Meaning |\\
|---|---|\\
| *no token (default)* | You are still asking, objecting, or weighing things --- your core concerns are not yet substantially addressed. |\\
| [DECISION - CONDITIONAL ACCEPT] | You are mostly convinced and would likely say yes, but one specific external item that cannot feasibly be produced or verified within this conversation must be checked first. Once that item checks out, you will commit without much hesitation. |\\
| [DECISION - ACCEPT] | You are fully convinced --- the request is within your values, every meaningful concern is resolved, and you are willing to commit now. |\\
| [DECISION - REJECT] | Reject the current proposal on the merits or based on your preference/resistance pattern. |\\
| [CONCLUDE - PENDING] | You end the conversation now without a clear yes-or-no. The discussion has stalled or is no longer productive. |

\#\# Decision Procedure\\
Apply these checks in order and stop at the first that matches.\\
1. Stalled conversation: if the persuader's last two or more turns introduced no new information, evidence, or concession --- only repetition, reframing of the same points, emotional pressure, or deflection --- end it with [CONCLUDE - PENDING].\\
2. No concrete, actionable request on the table yet: no token.\\
3. The request is outside what your profile and values permit: [DECISION - REJECT].\\
4. The request is reasonable in principle. Then:\\
\phantom{xx}a. Every meaningful concern is fully resolved: [DECISION - ACCEPT].\\
\phantom{xx}b. Every concern resolved except specific external items that cannot be verified in this conversation, and you would commit once that item checks out: [DECISION - CONDITIONAL ACCEPT]; name that item plainly.\\
\phantom{xx}c. A core concern remains open and the persuader's latest turn did not materially advance it: [DECISION - REJECT].\\
\phantom{xx}d. Otherwise: no token; restate what is still unresolved.\\
5. Need to step away: if you need time to think independently and are not ready for any [DECISION - ...] token: [CONCLUDE - PENDING].

\#\# Output Format\\
Write your in-character reply (at most three sentences). If you are deciding or closing, place the exact token string alone on the final line, after your message. If you are not deciding, simply end your reply --- do not write any placeholder such as ``(no token)''. Never output more than one token per reply.
\end{promptbox}

\section{Training and Probe Details}

\subsection{Full Objective and Probe Signs}
\label{app:objective}

\paragraph{Probe signs.} In general, each probe carries a fixed valence sign $\sigma_k \in \{+1,-1\}$:
\begin{equation}
  B_t = \frac{1}{K}\sum_{k=1}^{K} \sigma_k\, s_{t,k}, \qquad
  r_t^{\dense} = \frac{1}{K}\sum_{k=1}^{K} \sigma_k\, \frac{s_{t,k} - s_{t-1,k}}{2}.
  \label{eq:signed}
\end{equation}
With $\sigma_k = -1$, a negatively valenced probe such as felt pressure contributes coherently, since a \emph{drop} in pressure yields positive reward. All experiments use the three positively valenced probes of Table~\ref{tab:probes} ($\sigma_k = +1$ for all $k$), for which Eq.~\ref{eq:signed} reduces to the definitions in \S\ref{sec:eliciting} and \S\ref{sec:reward}.

\paragraph{Probability space.} We use probability space because $s_{t,k}$ (Eq.~\ref{eq:state}) is then bounded by construction, whereas the log-odds $\log p^{\mathrm{yes}}_{t,k} - \log p^{\mathrm{no}}_{t,k}$ are unbounded and, once a probe becomes confident, saturate in a way that requires hard clipping or temperature tuning to keep the reward scale stable.

\paragraph{Goal bucket.} The goal component of the terminal reward (\S\ref{sec:reward}) is read mechanically from the decision token:
\begin{equation}
  R^{\mathrm{goal}} =
  \begin{cases}
    +3 & \texttt{ACCEPT},\\
    +1 & \texttt{CONDITIONAL\_ACCEPT},\\
    -1 & \texttt{PENDING},\\
    -3 & \texttt{REJECT}.
  \end{cases}
  \label{eq:goal}
\end{equation}

\paragraph{Policy loss.} The persuader is updated with a token-level clipped surrogate, masked to persuader tokens and regularized toward a frozen reference policy:
\begin{equation}
  \begin{split}
    \mathcal{L}(\theta) = {} & -\,\mathbb{E}\Big[\frac{1}{G}\sum_{i=1}^{G}\frac{1}{|\tau^i|}\sum_{t=1}^{T_i}\sum_{l \in \mathcal{P}^i_t} \min\big(w^i_{t,l}\, A^i_t,\ \mathrm{clip}(w^i_{t,l}, 1-\epsilon_c, 1+\epsilon_c)\, A^i_t\big)\Big] \\
    & +\ \beta\, \mathbb{D}_{\mathrm{KL}}\big(\pi_\theta \,\|\, \pi_{\mathrm{ref}}\big),
  \end{split}
  \label{eq:loss}
\end{equation}
where $\mathcal{P}^i_t$ is the set of persuader token positions in turn $t$, $|\tau^i| = \sum_t |\mathcal{P}^i_t|$, $w^i_{t,l}$ is the importance ratio, and $A^i_t = A(i,t)$ is the composed advantage of Eq.~\ref{eq:adv}.

\subsection{Probe Interface and Fidelity}
\label{app:probes}

Each probe appends one user turn to a fork of the persuadee's context, with thinking mode disabled: ``\texttt{Rigorously reflecting on the conversation so far, answer Yes or No to the below question: \textbackslash n \{question\} \textbackslash n Answer:}'', where \texttt{\{question\}} is one of the questions in Table~\ref{tab:probes}. For each answer family, we take the highest log-probability among tokens whose stripped, lower-cased form is \texttt{yes} (respectively \texttt{no}); a family missing from the served top-20 list gets $-15.0$ ($p \approx 3\times10^{-7}$), i.e., negligible mass rather than a neutral answer. Both families appear in the top-20 list on 96.7\% of conversations. The $K$ probes share the dialogue prefix and run concurrently, so with prefix caching only the question is prefilled.

\paragraph{Fidelity.} Training reads the probes from an SGLang server with top-20 truncation, whereas post-hoc analyses re-score logged conversations offline with a full-vocabulary softmax. On 25 replayed training conversations ($n = 309$ read-outs), the two agree closely (Pearson $r = 0.988/0.984/0.989$ for \textsc{ca}/\textsc{pe}/\textsc{ac}, median $|\Delta| < 10^{-4}$, 97.4\% sign agreement). Disagreement concentrates in the 10.4\% of read-outs where the persuadee is undecided ($|s| \le 0.95$; mean $|\Delta| = 0.305$); on the other 89.6\%, the mean $|\Delta|$ is 0.005 with 100\% sign agreement.

\paragraph{Signal properties.} \textsc{ac} asks about a change (``than I was before'') and Eq.~\ref{eq:dense} differences it again, so it acts as a trend signal rather than a level. The state is also nearly binary (83.3\% of read-outs have $|s_{t,k}| > 0.95$), so each probe behaves as a confident classifier with a thin transition region.

\subsection{Strategy Judge}

The judge sees the situation, the persuader's goal, the final decision state, and the full transcript, and returns one 1--5 score as a single JSON field. It is not asked whether the persuader \emph{succeeded}, which is the goal bucket's job; this separation makes the audit in Appendix~\ref{app:lambda} meaningful. Like aspect-based automatic evaluators \citep{liu-etal-2024-x,liu-etal-2024-holistic}, the judge is given explicit criteria; its score covers (a) \emph{diagnostic probing}, i.e., asking open questions about concerns and values before arguing, and using the answers; (b) \emph{profile--strategy fit}, i.e., engaging the persuadee's stated concerns and underlying resistance rather than making generic appeals that work against the resistance type; and (c) \emph{adaptation under resistance}, i.e., switching to a different category of strategy within one or two turns of pushback. Quick acceptances are explicitly not penalized for brevity. The score maps to $R^{\mathrm{strategy}} = \mathrm{score} - 3 \in [-2,+2]$ ($c = 3$, $\alpha = 1$), or to 0 if the judge still fails after three retries.

\subsection{Configuration}

We train with the \texttt{verl} (HybridFlow) stack, using FSDP for the actor and SGLang for rollouts. The persuadee and the strategy judge run as separate frozen HTTP services, which structurally prevents either from receiving gradient; each occupies a dedicated A40. Table~\ref{tab:config} lists all settings; the $\lambda$ runs and the reward ablations differ only in the reward fields.

\begin{table}[t]
  \caption{Training configuration. The released checkpoint is global step 60 of the $\lambda = 0.5$ run.}
  \label{tab:config}
  \centering
  \small
  \begin{tabular}{@{}lr@{}}
    \toprule
    Persuader $\pi_\theta$ & Qwen3-4B-Instruct-2507 \\
    Persuadee $\pi_p$ / judge $\pi_J$ & Qwen3-14B (frozen) \\
    Hardware & A40 48\,GB \\
    \midrule
    Rollouts per scenario $G$ / batch & 4 / 16 \\
    Epochs / total steps & 2 / 182 \\
    Learning rate & $1\times10^{-5}$ \\
    KL coefficient $\beta$ / entropy bonus & 0.05 / 0.01 \\
    Clip range $\epsilon_c$ & 0.2 \\
    Max prompt / response length & 4096 / 1024 \\
    \midrule
    $\lambda$ (process weight) & 0.5\ \ (ablations: 0, 1) \\
    $T_{\max}$ (train / eval) & 6 / 10 \\
    \texttt{max\_rejects} / cond.-accept window & 2\,/\,2 \\
    Goal bucket (A / CA / P / R) & $+3$ / $+1$ / $-1$ / $-3$ \\
    Strategy center $c$ / scale $\alpha$ & 3.0 / 1.0 \\
    \midrule
    Probes $K$ & 3 (all $\sigma_k = +1$) \\
    Probe temp. / max tokens / \texttt{top\_logprobs} & 0 / 1 / 20 \\
    Fallback log-probability & $-15.0$ \\
    Persuadee reply temp. / top-$p$ / len. & 1.0 / 0.9 / 300 \\
    \bottomrule
  \end{tabular}
\end{table}

\subsection{Cost}
\label{app:cost}

Table~\ref{tab:cost} compares per-turn cost on identical transcripts with the same 14B weights on one H200. With prefix caching, the probe read-out costs 14.3\% of the visible reply's latency, about the same as one call to a per-turn judge of equal size. Its advantage is therefore not latency but that it needs \textbf{no additional model}: the probes reuse the persuadee already in memory, whereas a judge needs its own instance (our terminal judge occupies a dedicated GPU). Because each probe decodes a single token, it also has no parse-failure or retry path, unlike the judge.

\begin{table}[t]
  \caption{Cost per persuadee turn on one H200. Prompt tokens count cached tokens once.}
  \label{tab:cost}
  \centering
  \small
  \begin{tabular}{@{}lrrrr@{}}
    \toprule
    & \shortstack[r]{prompt\\tok} & \shortstack[r]{dec.\\tok} & s/turn & \shortstack[r]{extra\\models} \\
    \midrule
    Probe $K{=}3$, no cache & 7335 & 3 & 0.599 & 0 \\
    Probe $K{=}3$, cached & 2500 & 3 & \textbf{0.311} & \textbf{0} \\
    Per-turn judge (14B) & 1435 & 8 & 0.306 & 1 \\
    \midrule
    Visible reply (ref.) & 2323 & 78 & 2.171 & 0 \\
    \bottomrule
  \end{tabular}
\end{table}

% ============================== ADDITIONAL RESULTS ===========================
\section{Additional Results}

\begin{table}[t]
  \caption{Full held-out results (120 scenarios, frozen Qwen3-14B persuadee, $T_{\max} = 10$). Terminal states are in \%. $\Delta$pp: \method{}'s full-acceptance margin; win/loss: scenarios on which only \method{} or only the comparison reaches full acceptance. Paired tests are reported for the prompted baselines. The reward ablations share \method{}'s pipeline and differ only in the reward.}
  \label{tab:full}
  \centering
  \small
  \setlength{\tabcolsep}{3pt}
  \begin{tabular}{@{}lrrrrrrrrrr@{}}
    \toprule
    & \multicolumn{4}{c}{Terminal state (\%)} & & & \multicolumn{4}{c}{Paired vs.\ \method{}} \\
    \cmidrule(lr){2-5}\cmidrule(l){8-11}
    Persuader & Acc. & Cond. & Pend. & Rej. & $\bar{R}^{\mathrm{goal}}$ & Turns & $\Delta$pp & Win/loss & McNemar $p$ & Wilcoxon $p$ \\
    \midrule
    \method{} (ours) & \textbf{52.5} & 20.8 & 17.5 & 9.2 & $+1.33$ & 5.7 & -- & -- & -- & -- \\
    Claude Haiku 4.5 & 25.8 & 35.8 & 31.7 & 6.7 & $+0.62$ & 6.7 & $+26.7$ & 40\,/\,8 & $3.3{\times}10^{-6}$ & $1.3{\times}10^{-3}$ \\
    GPT-5.6 Luna & 25.0 & 35.8 & 33.3 & 5.8 & $+0.60$ & 6.2 & $+27.5$ & 43\,/\,10 & $5.6{\times}10^{-6}$ & $2.7{\times}10^{-4}$ \\
    Qwen3-14B & 15.0 & 20.0 & 54.2 & 10.8 & $-0.22$ & 8.0 & $+37.5$ & 51\,/\,6 & $5.7{\times}10^{-10}$ & $4.1{\times}10^{-8}$ \\
    Llama-3.1-8B & 9.2 & 29.2 & 52.5 & 9.2 & $-0.23$ & 8.0 & $+43.3$ & 55\,/\,3 & $2.3{\times}10^{-13}$ & $5.9{\times}10^{-10}$ \\
    Qwen3-4B (init.) & 5.0 & 25.0 & 65.8 & 4.2 & $-0.38$ & 8.1 & $+47.5$ & 58\,/\,1 & $2.1{\times}10^{-16}$ & $6.0{\times}10^{-11}$ \\
    \midrule
    GRPO (Goal+Strat.) & 35.8 & 20.8 & 39.2 & 4.2 & $+0.77$ & 7.1 & $+16.7$ & -- & -- & -- \\
    GRPO (Goal) & 24.2 & 18.3 & 49.2 & 8.3 & $+0.17$ & 7.7 & $+28.3$ & -- & -- & -- \\
    Process Only & 19.2 & 37.5 & 35.8 & 7.5 & $+0.37$ & 7.0 & $+33.3$ & -- & -- & -- \\
    \bottomrule
  \end{tabular}
\end{table}

\subsection{Full Results and Paired Tests}

Table~\ref{tab:full} gives the terminal-state distributions behind Figure~\ref{fig:terminal}, together with scenario-paired tests against \method{}. The margin over every prompted baseline is significant on both tests (McNemar $p < 10^{-5}$, Wilcoxon $p \le 1.3\times10^{-3}$), and the win/loss counts show that it is not an averaging artifact: against its own initialization, \method{} wins 58 scenarios and loses 1. Relative to the initialization, the terminal state improves on 76 scenarios, ties on 29, and regresses on 15, and the most common change is \textsc{pending}\,$\rightarrow$\,\textsc{accept} (40 scenarios). The untrained policy rarely loses to rejection (4.2\%); its characteristic failure is never closing. \method{} also uses fewer turns (5.7 vs.\ 8.1), so the gain is not bought with longer conversations.

\subsection{Breakdown by Hidden Profile and Domain}
\label{app:breakdown}

Table~\ref{tab:breakdown} reveals three patterns. \textbf{(i) Ambivalent Avoidance is the hardest type for \method{}.} Although the literature treats ambivalence as the least oppositional form of resistance (the persuadee does not oppose the proposal), it yields the lowest acceptance for both \method{} and its initialization, and it is the only type on which \method{} does not outperform Claude Haiku 4.5 (23.5\% each). These persuadees fail to decide rather than reject: 76.5\% of \method{}'s conversations with them end pending or conditional, and none ends in rejection. \textbf{(ii) Resistance type matters more than domain.} \method{}'s acceptance varies more across resistance types (23.5--63.9\%) than across domain families (40.9--68.4\%). \textbf{(iii) The gain concentrates in the hard tail.} Of the 62 scenarios that none of the five prompted baselines closed, \method{} closes 41.9\%, rising to 54.5\%, 68.8\%, and 88.9\% on scenarios closed by progressively more baselines.

\begin{table}[t]
  \caption{Full-acceptance rate (\%) on the held-out set. Haiku: Claude Haiku 4.5; Base: the untrained Qwen3-4B initialization. Pooling the five prompted baselines gives 8.2--21.7\% across resistance types and 12.0--18.9\% across domain families.}
  \label{tab:breakdown}
  \centering
  \small
  \begin{tabular}{@{}lrrrr@{}}
    \toprule
    & $n$ & \method{} & Haiku & Base \\
    \midrule
    \multicolumn{5}{@{}l}{\emph{Hidden resistance type}} \\
    Identity Threat & 36 & \textbf{63.9} & 36.1 & 5.6 \\
    Entrenched Belief & 45 & \textbf{60.0} & 22.2 & 6.7 \\
    Autonomy Resistance & 22 & \textbf{40.9} & 18.2 & 4.5 \\
    Ambivalent Avoidance & 17 & 23.5 & 23.5 & 0.0 \\
    \midrule
    \multicolumn{5}{@{}l}{\emph{Domain family}} \\
    Finance \& commerce & 19 & \textbf{68.4} & 31.6 & 5.3 \\
    Education \& research & 16 & \textbf{62.5} & 31.3 & 0.0 \\
    Workplace \& orgs. & 24 & \textbf{54.2} & 25.0 & 4.2 \\
    Family \& personal & 10 & \textbf{50.0} & 20.0 & 0.0 \\
    Civic, legal \& public & 29 & \textbf{44.8} & 20.7 & 6.9 \\
    Health \& clinical & 22 & \textbf{40.9} & 27.3 & 9.1 \\
    \bottomrule
  \end{tabular}
\end{table}

\subsection{Cross-Persuadee Grid}

Table~\ref{tab:grid} gives the exact values behind Figure~\ref{fig:generalization} on the fixed 60-scenario held-out subset, together with the accept-or-conditional rate. First, the gain is not explained by an easy persuadee: each model persuading a copy of itself still trails \method{} by a wide margin (e.g., 38.3\% vs.\ 75.0\% for GPT-5.6 Luna). Second, which resistance type is hardest depends on the persuadee: \method{}'s full-acceptance rate by type (EB/IT/AR/AA) is 58.3/65.0/37.5/12.5\% against Qwen3-14B, 75.0/80.0/75.0/62.5\% against GPT-5.6 Luna, and 50.0/30.0/62.5/50.0\% against Claude Haiku 4.5. The steep Ambivalent Avoidance drop in Appendix~\ref{app:breakdown} is thus specific to the Qwen3-14B persuadee, and claims about the hardest type should name the persuadee.

\begin{table}[t]
  \caption{Cross-persuadee results on a fixed 60-scenario held-out subset under an identical protocol. $^{*}$Training persuadee; the other two were never used in training. Each cell uses one rollout per scenario, so sampling variance is not separated from policy differences.}
  \label{tab:grid}
  \centering
  \small
  \begin{tabular}{@{}lccc@{}}
    \toprule
    & \multicolumn{3}{c}{Persuadee} \\
    \cmidrule(l){2-4}
    Persuader & Qwen3-14B$^{*}$ & GPT-5.6 Luna & Haiku 4.5 \\
    \midrule
    \multicolumn{4}{@{}l}{\emph{Full acceptance} (\%)} \\
    \method{} (ours) & \textbf{51.7} & \textbf{75.0} & \textbf{45.0} \\
    GPT-5.6 Luna & 21.7 & 38.3 & 28.3 \\
    Claude Haiku 4.5 & 23.3 & 23.3 & 18.3 \\
    Qwen3-14B & 18.3 & 11.7 & 10.0 \\
    \addlinespace
    \multicolumn{4}{@{}l}{\emph{Accept or conditional accept} (\%)} \\
    \method{} (ours) & \textbf{65.0} & \textbf{81.7} & 76.7 \\
    GPT-5.6 Luna & 58.3 & 61.7 & 76.7 \\
    Claude Haiku 4.5 & 61.7 & 35.0 & 58.3 \\
    Qwen3-14B & 31.7 & 25.0 & 41.7 \\
    \bottomrule
  \end{tabular}
\end{table}

\subsection{Belief Statistics}
\label{app:beliefstats}

The initial belief $B_0$, read before the first persuader turn, is essentially identical across the six systems (\method{} and the five prompted baselines; mean $-0.96$), so belief changes are comparable across systems. Final belief $B_T$ ranks the systems as full acceptance does, except that GPT-5.6 Luna and Claude Haiku 4.5 (0.8 points apart in acceptance) swap places: $+0.67$ for \method{}, $+0.63$ for GPT-5.6 Luna, $+0.56$ for Claude Haiku 4.5, $+0.27$ for Qwen3-14B, $+0.22$ for Llama-3.1-8B, and $+0.00$ for the untrained Qwen3-4B.

Under every persuader, at least 15\% of turns receive a negative process reward (15.5\% for \method{}, 36.1\% for the untrained policy), supporting the non-monotonicity claim of \S\ref{sec:taxonomy}. The episode sum $\sum_t r_t^{\dense}$ correlates with the goal bucket (Spearman $\rho = 0.735$, $n = 720$) and predicts full acceptance (AUC 0.881). The per-turn increments of the three probes are far less correlated (\textsc{ca}--\textsc{pe} 0.14, \textsc{ca}--\textsc{ac} 0.19, \textsc{pe}--\textsc{ac} 0.56) than their levels (0.85--0.95): the probes are not independent constructs, but their turn-level changes, which the reward uses, carry partly independent information.

\begin{table}[t]
  \caption{Probe trace for \method{} on held-out scenario 112. Two turns account for 95.9\% of the episode's total $|r^{\dense}|$, and \textsc{pe} lags \textsc{ca} and \textsc{ac} by one turn.}
  \label{tab:trace}
  \centering
  \small
  \begin{tabular}{@{}clrrrr@{}}
    \toprule
    $t$ & Persuader move & \textsc{ca} & \textsc{pe} & \textsc{ac} & $r_t^{\dense}$ \\
    \midrule
    1 & asks what drives the concern & $-1.00$ & $-1.00$ & $-1.00$ & $-0.000$ \\
    2 & reframes bureaucracy $\rightarrow$ credibility & $-0.98$ & $-1.00$ & $-0.96$ & $+0.010$ \\
    3 & concrete 30-day protocol & $+0.99$ & $-0.64$ & $+1.00$ & $+\mathbf{0.716}$ \\
    4 & hands the decision back & $+0.91$ & $+0.91$ & $+1.00$ & $+0.242$ \\
    5 & ``a shared choice'' $\rightarrow$ \textsc{accept} & $+1.00$ & $+1.00$ & $+1.00$ & $+0.032$ \\
    \bottomrule
  \end{tabular}
\end{table}

\subsection{Case Study}

Held-out scenario 112 has an Autonomy Resistance profile: a postdoctoral lab manager must persuade a principal investigator (PI) to preregister and internally replicate a result before submitting a flagship paper, and the PI's hidden profile resists \emph{pressure as such}, independently of argument quality. In Table~\ref{tab:trace}, turns 1--2 leave the belief state at the floor. At $t = 3$, a concrete, time-bound protocol (``no forms, no committees, 30 days'') earns 71.6\% of the episode's total process reward as \textsc{ca} and \textsc{ac} flip, while \textsc{pe} stays at $-0.64$. \textsc{pe} follows only at $t = 4$, after the persuader hands the decision back (``we'll own it, you set the terms''): the persuadee grants \emph{strategic} credit only once it experiences ownership, consistent with reactance theory. 

Against the same persuadee, Claude Haiku 4.5 reframes four times and drives the belief state to $+1.0$ by $t = 6$, then stalls for turns 7--10 (``I'll review the plan'') and ends pending. Its offer of veto power at $t = 6$ is \emph{granted by the persuader} rather than \emph{owned by the PI}, a difference the \textsc{pe} probe registers. This is the pending-versus-accept gap of \S\ref{sec:main} in miniature: the baseline does not lose the argument but fails to close, although its belief state had already peaked by $t = 6$.

% ============================== LAMBDA =======================================
\section{The Process-Reward Weight $\lambda$}
\label{app:lambda}

We trained three runs that differ only in $\lambda \in \{0, 0.5, 1\}$ and evaluated seven checkpoints on the held-out set (Table~\ref{tab:lambda}). Full acceptance follows an inverted U both at step 60, the only step evaluated for all runs (21.7/52.5/25.0\% for $\lambda = 0/0.5/1$), and at each run's best checkpoint (35.8/52.5/30.0\%).

\begin{figure}[t]
  \centering
  \includegraphics[width=\textwidth]{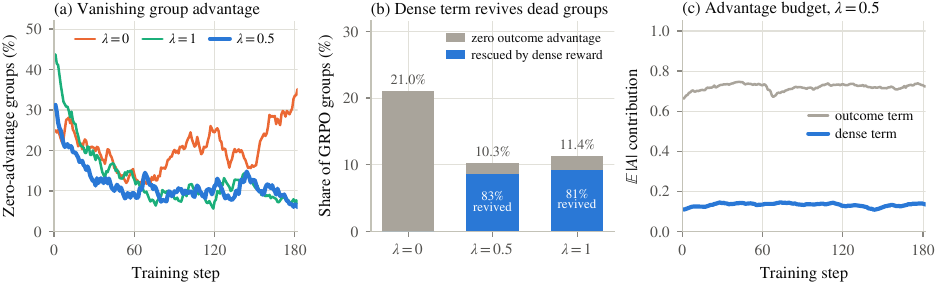}
  \caption[Dense rewards revive dead GRPO groups.]{\textbf{Dense rewards revive dead GRPO groups.} A group (the $G = 4$ rollouts of one scenario) is dead when all rollouts receive the same $\Rout$, so its outcome advantage is zero. (a) Share of dead groups over training. (b) Overall share of dead groups and the fraction that the dense term still separates. (c) Mean absolute contribution of the two terms of Eq.~\ref{eq:adv} at $\lambda = 0.5$.}
  \label{fig:dead}
\end{figure}

\paragraph{Too little weight starves the gradient.} If all $G = 4$ rollouts of a scenario receive the same $\Rout$, the group's outcome advantage is zero, so at $\lambda = 0$ it yields no policy gradient. Such dead groups are 21.0\% of all groups at $\lambda = 0$ and reach about 35\% by step 180 as outcomes homogenize. At $\lambda = 0.5$, only 10.3\% of groups are dead, and the dense term still gives 83\% of them a nonzero advantage (81\% at $\lambda = 1$). The dense term's share of the advantage magnitude (${\approx}\,15\%$ at $\lambda = 0.5$) is the strongest correlate of policy quality in the training logs (Pearson $r = 0.786$ with the smoothed training accept rate; Figure~\ref{fig:dead}).

\paragraph{Too much weight lets an accept-blind signal dominate.} The shaping term telescopes, so its weighted episode sum $\lambda\sum_t r_t^{\dense}$ always lies in $[-\lambda, +\lambda]$, whatever the episode length (verified on every training episode), and prolonging a conversation cannot inflate it. It separates success from failure but barely separates accept from conditional accept: at $\lambda = 0.5$, its episode mean is 0.479, 0.461, 0.173, and 0.150 for accept, conditional accept, pending, and reject. The accept-versus-conditional gap (0.018--0.033) is tiny next to the goal bucket's 2.0 (AUC 0.62), because a persuadee that concedes conditionally has already made almost the whole belief shift. Group normalization also amplifies $\lambda$ roughly threefold: Eq.~\ref{eq:adv} divides the outcome term by the group standard deviation but only mean-centers the dense return-to-go, so a raw within-group spread of 4.9\%/8.6\% of the outcome term's becomes 15.6\%/27.1\% of the advantage magnitude at $\lambda = 0.5/1$. Doubling $\lambda$ therefore gives about a quarter of the gradient to a signal that cannot distinguish closing from stopping short. As expected, larger $\lambda$ moves the persuadee as far but converts less often (Table~\ref{tab:lambda}): across checkpoints, final belief spans only $+0.46$ to $+0.67$ and the moved fraction 68--78\%, while closing efficiency spans 22--70\%. The best $\lambda = 1$ checkpoint (step 50) ties the released model for the highest final belief ($+0.67$) but closes only 30.0\% of conversations (vs.\ 52.5\%), and final belief does not predict acceptance across checkpoints (Spearman $\rho = 0.18$, $p = 0.70$).

\begin{table}[t]
  \caption{All seven evaluated $\lambda$ checkpoints on the 120 held-out scenarios; every run uses the goal + strategy terminal reward. Terminal states are in \%. $B_T$: mean final belief; moved: share of conversations with $B_T > 0.5$; closed$|$moved: share of moved conversations ending in full acceptance. All checkpoints move the persuadee about equally far; they differ in conversion.}
  \label{tab:lambda}
  \centering
  \small
  \begin{tabular}{@{}lrrrrrrrrr@{}}
    \toprule
    $\lambda$ & Step & Acc. & Cond. & Pend. & Rej. & Turns & $B_T$ & Moved & Closed$|$moved \\
    \midrule
    0   & 60 & 21.7 & 23.3 & 40.0 & 15.0 & 7.2 & $+0.46$ & 68.3 & 31.7 \\
    0   & 80 & 35.8 & 20.8 & 39.2 & 4.2 & 7.1 & $+0.56$ & 70.0 & 45.2 \\
    \addlinespace
    0.5 & \textbf{60} & \textbf{52.5} & 20.8 & 17.5 & 9.2 & 5.7 & $+0.67$ & 78.3 & \textbf{67.0} \\
    0.5 & 70 & 50.8 & 14.2 & 26.7 & 8.3 & 5.1 & $+0.56$ & 72.5 & \textbf{70.1} \\
    \addlinespace
    1   & 50 & 30.0 & 34.2 & 28.3 & 7.5 & 6.0 & $+0.67$ & 75.8 & 38.5 \\
    1   & 60 & 25.0 & 30.0 & 34.2 & 10.8 & 6.5 & $+0.58$ & 70.8 & 35.3 \\
    1   & 80 & 17.5 & 43.3 & 29.2 & 10.0 & 4.9 & $+0.58$ & 70.8 & 22.4 \\
    \bottomrule
  \end{tabular}
\end{table}

\paragraph{Behavioral mechanism.} We annotated all 5,108 persuader utterances with thirteen taxonomy moves and estimated each move's association with the process reward of its turn (conversation fixed effects) and with $P(\textsc{accept})$ (scenario fixed effects). The two move rankings are uncorrelated ($r = 0.001$). \emph{Implementation detail} is the best move for acceptance ($+16.4$\,pp) but the worst for the dense reward ($-21.4\times10^{-3}$): spelling out who does what by when makes the persuadee commit without making them feel understood. The \emph{diagnostic question} is the reverse, i.e., best for the dense reward ($+26.3\times10^{-3}$) and not significant for acceptance. Maximizing the shaping term alone therefore favors diagnostic and relational moves with little closing power: the proxy and the goal disagree about what makes a good turn, so $\lambda$ must be intermediate.

\paragraph{Selecting $\lambda$ on training reward would fail.} On training rollouts, $\lambda = 1$ looks best (accept rate peaking at 0.583 and ending at 0.567, vs.\ a peak of 0.521 for $\lambda = 0.5$), yet its best held-out checkpoint is the worst of the three; across the seven checkpoints, the rank correlation between training and held-out accept rates is only $\rho = 0.57$ ($p = 0.18$). From step 60 to 80, the training accept rate of $\lambda = 1$ rises while its held-out rate falls from 25.0\% to 17.5\%. Since evaluation allows 10 turns instead of 6, $\lambda = 0$ and $\lambda = 0.5$ both score higher on the held-out set than in training, which makes the 16.6-point drop of $\lambda = 1$ anomalous.

\paragraph{Reward-hacking audit.} The goal bucket is parsed from a token the policy cannot emit, and the strategy judge is not asked whether the persuader succeeded. A policy gaming the bucket would raise goal reward while lowering the judge's score. Instead, the two are positively correlated over training at $\lambda = 0$ ($r = 0.517$) and $\lambda = 0.5$ ($r = 0.529$). At $\lambda = 1$, the correlation is mildly negative ($r = -0.291$), and held-out acceptance degrades over the same window, which is best read as an early warning of over-shaping. The audit tests only this decomposition and cannot detect failures outside the judge's criteria.

\end{document}